\def\year{2022}\relax
\documentclass[letterpaper]{article} 
\usepackage{aaai22}  
\usepackage{times}  
\usepackage{helvet}  
\usepackage{courier}  
\usepackage[hyphens]{url}  
\usepackage{graphicx} 
\usepackage{natbib}  
\usepackage{caption} 
\DeclareCaptionStyle{ruled}{labelfont=normalfont,labelsep=colon,strut=off} 
\usepackage{algorithm}
\usepackage{algorithmic}
\usepackage{caption}
\usepackage{subfigure}
\usepackage{newfloat}
\usepackage{listings}
\floatstyle{ruled}
\newfloat{listing}{tb}{lst}{}
\floatname{listing}{Listing}
\usepackage{amsmath,amsfonts,bm}
\usepackage{amssymb}
\usepackage{booktabs}
\usepackage{multirow}

\title{AdaPruner: Adaptive Channel Pruning and Effective Weights Inheritance}
\author {
    Xiangcheng Liu, Jian Cao, Hongyi Yao, Wenyu Sun, Yuan Zhang
}
\affiliations{
    School of Software and Microelectronics, Peking University\\
    \{liuxiangcheng, yhy\}@stu.pku.edu.cn, caojian@ss.pku.edu.cn,\\
    sunwenyu@pku.edu.cn, gump\_well\_done@163.com
}

\begin{document}

\maketitle

\begin{abstract}
Channel pruning is one of the major compression approaches for deep neural networks. While previous pruning methods have mostly focused on identifying unimportant channels, channel pruning is considered as a special case of neural architecture search in recent years. However, existing methods are either complicated or prone to sub-optimal pruning. In this paper, we propose a pruning framework that adaptively determines the number of each layer's channels as well as the wights inheritance criteria for sub-network. Firstly, evaluate the importance of each block in the network based on the mean of the scaling parameters of the BN layers. Secondly, use the bisection method to quickly find the compact sub-network satisfying the budget. Finally, adaptively and efficiently choose the weight inheritance criterion that fits the current architecture and fine-tune the pruned network to recover performance. AdaPruner allows to obtain pruned network quickly, accurately and efficiently, taking into account both the structure and initialization weights. We prune the currently popular CNN models (VGG, ResNet, MobileNetV2) on different image classification datasets, and the experimental results demonstrate the effectiveness of our proposed method. On ImageNet, we reduce 32.8\% FLOPs of MobileNetV2 with only 0.62\% decrease for top-1 accuracy, which exceeds all previous state-of-the-art channel pruning methods. The code will be released.
\end{abstract}

\section{Introduction}
\noindent The widespread application of convolutional neural networks \cite{szegedy2015going,ren2015faster,long2015fully} in computer vision is a great success. However, deep neural networks, while achieving high performance, are also suffering from problems such as oversized storage space and large computation, which hinder the deployment and application of CNNs in edge devices.  Channel pruning is an effective model compression technique that reduces the width of the model by pruning some channels in the network, significantly compressing the storage space of weights and speeding up the inference calculation.

Traditional pruning methods usually follow the following procedure: Train a large over-parameterized network. Prune the unimportant channels in each layer according to various predefined criteria. Maintain the weights of the original network and retrain it. These criteria include the L1 norm of the convolution kernel \cite{li2016pruning}, the percentage of zeros among the output activation \cite{hu2016network} and the reconstruction error for the next layer \cite{luo2017thinet}.  The last two steps are often iterated in order to achieve higher pruning rate and better performance.

A recent work \cite{liu2018rethinking} suggests that the pruned network architecture itself may be more important than the weights inherited from the original network, which inspired researchers to find more efficient compact sub-networks. Several pruning methods based on automatic machine learning have been developed. AMC \cite{he2018amc} proposes to train an intelligent agent that outputs the pruning rate of each layer by reinforcement learning. MetaPruning \cite{liu2019metapruning} uses meta-learning to train a PruningNet, generating specific weights for sub-networks with different structures, and later searching for the optimal sub-network by an evolutionary algorithm. ABCPruner \cite{lin2020channel} automatically searches for efficient network architectures with artificial bee colony algorithm and fine-tunes them to select the most efficient sub-network. However, these methods have complicated processes and require iterative searches that face high computational costs and may lead to sub-optimal search results.

\begin{figure*}[t]
\centering
\includegraphics[width=1.0\textwidth]{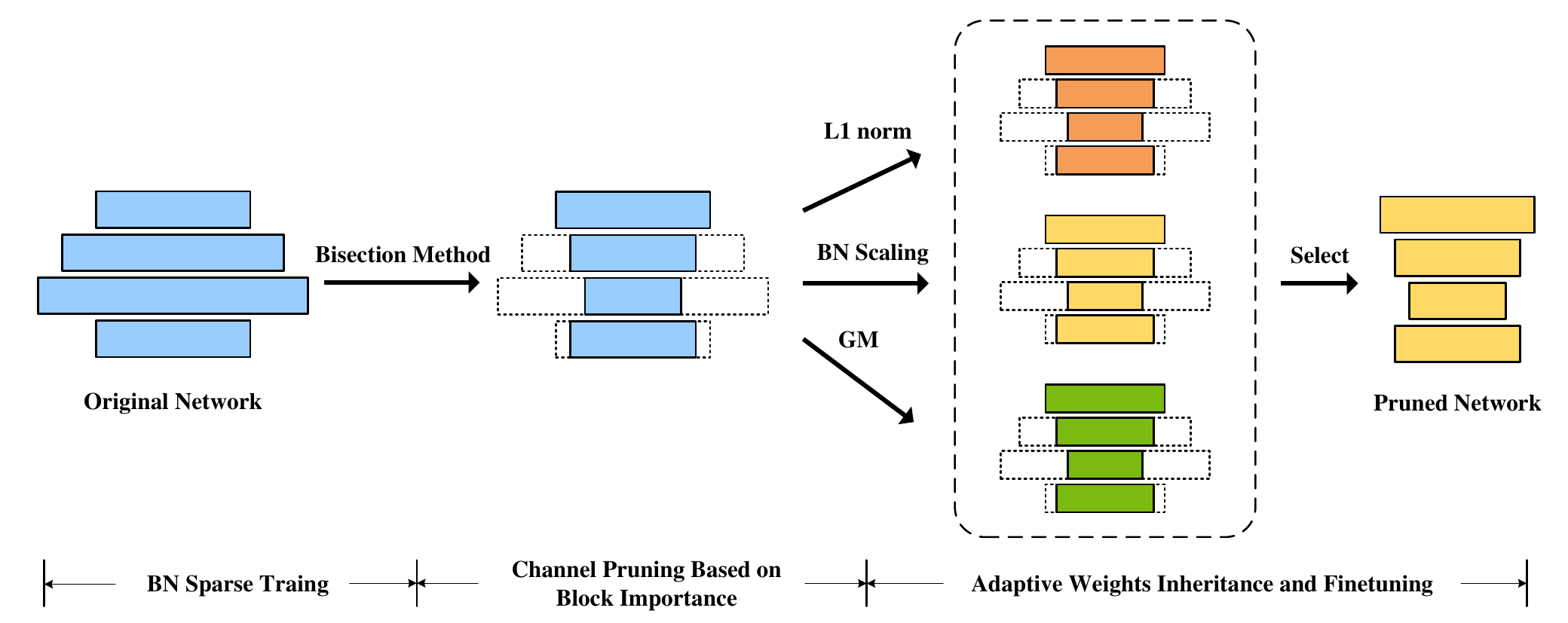} 
\caption{Framework of AdaPruner. Firstly, regularization is added to the scaling factors of the BN layer to encourage sparsity, and the importance of the block is estimated from the mean of the scaling factors. Secondly, we obtain the pruning rate of each block based on its importance, and this process is accomplished quickly with Bisection method. Finally, adaptive weights inheritance is performed on the sub-network meeting the budget to evaluate the optimal inheritance and finetune to recovery accuracy.}
\label{fig1}
\end{figure*}

In this paper, we propose a simple and efficient channel pruning method named AdaPruner. AdaPruner follows the three-step pipeline of traditional pruning methods, but integrates the structure and initialization weights of the pruned network, as shown in Figure \ref{fig1}. Firstly, sparse training is performed by adding L1 regularization to the gamma parameters of the BN layer, which allows us to evaluate the importance of different blocks based on the mean of all gamma parameters of their BN layers. Subsequently, we use bisection method to find the corresponding pruning rate of each block based on its respective importance, and thus obtain a compact sub-network conforming to the budget constraint. Finally, adaptively select the criterion to inherit the original network weights by recalibrating the BN layer parameters and fine-tune to recover the performance. AdaPruner obtains the efficient sub-network with only one BN sparse training, and adaptively selecting the inheritance criterion does not require retraining the pruned network, therefore our approach is simple and fast enough. We conduct experiments on mainstream CNN models on image classification benchmark CIFAR-10 and large-scale ImageNet datasets. The results demonstrate the effectiveness and superiority of our proposed method of adaptive channel number search combined with adaptive weight inheritance for network pruning.


\section{Related Work}
Model pruning techniques aim to remove redundant parameters from deep neural networks, thus achieving the goal of reducing model storage space and accelerating network inference speed. There are two broad categories of model pruning, structured pruning and \textbf{unstructured pruning}. The unstructured pruning technique \cite{lecun1990optimal,hassibi1993optimal,han2015deep,han2015learning} prunes the network at the fine-grained level so that the unimportant weight parameters become 0, and the final weights of the pruned network are sparse matrices. The advantage of unstructured pruning technique is that it maintains high accuracy even at large compression rate, while the weakness is that it requires specialized sparse matrix computation libraries and hardware support, which is not conducive to edge device deployment.

Structured pruning, also known as \textbf{channel pruning}, directly removes the entire convolutional filter, thus making the output channels of a particular layer fewer and reducing the width of the network. Channel pruning does not require special hardware or library support and allows for convenient inference acceleration, so there have been many research efforts focusing on structured pruning. The early work identifies the unimportant channels by various predefined criteria. For example, in \cite{li2016pruning}, the authors consider that the convolutional kernels with small L1 norm are less important than the others and can be pruned. Network Slimming \cite{liu2017learning} adds constraints to make the scaling factors of Batch Normalization layer sparse during training, and then prunes the convolutional kernels corresponding to small scaling factors. FPGM \cite{he2019filter} calculates the geometric median of the convolutional filters in the same layer, and those filters closest to it are pruned. ThiNet \cite{luo2017thinet} guides the pruning of the current layer based on the statistical information of the next layer, and minimizes the error of the next layer reconstruction in accordance with a greedy strategy. HRank \cite{lin2020hrank} performs feature decomposition on the output features and prune filters with low-rank feature maps. Traditional pruning methods treat different layers equally or artificially specify pruning rates for each layer, and then prune unimportant output channels, which leads to their sub-optimal pruning.

A recent work \cite{liu2018rethinking} has experimentally demonstrated that the pruned network architecture itself is more important than the inherited weights. Channel pruning can be considered as a special case of network architecture search (NAS), where only the output channel numbers of each layer are searched. Therefore a lot of \textbf{AutoML} based pruning methods have been proposed recently. AMC \cite{he2018amc} takes the compression rate and accuracy as feedback and uses reinforcement learning to obtain an intelligent agent yielding the pruning rate of each layer. However, there is usually an unstable convergence of reinforcement learning, which requires much effort to tune the parameters. MetaPruning \cite{liu2019metapruning} randomly samples each layer's output channel number as input, trains a PruningNet to generate high-quality weights for sub-networks of different architectures, and finally uses an evolutionary algorithm to search for the optimal sub-network that satisfies the constraints. EagleEye \cite{li2020eagleeye} proposes a way to quickly evaluate the accuracies of sub-networks and randomly generate a large number of sub-networks to select the best pruned network architecture. ABCPruner \cite{lin2020channel} utilizes an artificial bee colony algorithm \cite{karaboga2005idea} to directly search for the channel number configuration of the model, nevertheless it requires retraining the sub-network for performance evaluation, which results in a high computational cost.

\begin{figure}[t]
\centering
\includegraphics[width=0.9\columnwidth]{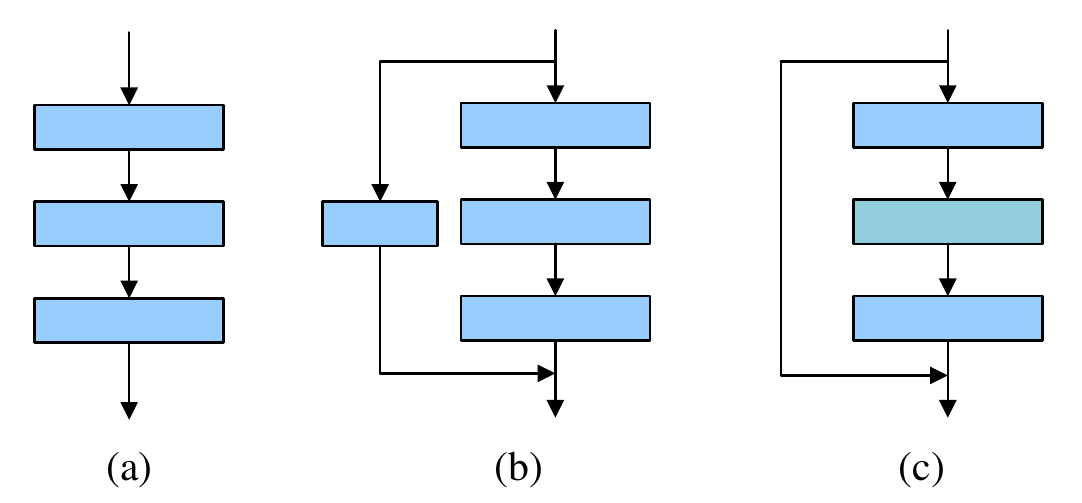} 
\caption{Illustration of the basic unit of channel pruning for different network architectures. (a) VGG16. For straight and unbranched networks, we perform channel pruning at each layer. (b) ResNet50. As the figure shows one of the residual blocks in ResNet50, we use it as the unit (Block) for pruning. (c) MobileNetV2. Inverted residual block is its constituent unit, and we also regard it as the block required for pruning. The middle layer of the inverted residual block is Depthwise Convolution.}
\label{fig2}
\end{figure}

\section{Methodology}
In this section, we elaborate the proposed AdaPruner method, and we propose to perform channel pruning in the following three steps. First, sparsely train the network and estimate the importance of each block to be pruned in the whole network based on its scaling factors mean value. Then, using the bisection method, determine the corresponding pruning rate of each block according to its importance. Finally, different weight inheritance criteria are quickly evaluated to determine the best criterion and retrain to recover the accuracy.

\subsection{Definition of Channel Pruning}
Given a trained CNN model containing \emph{n} layers with weights $\bm{W}$, the output channel number of each layer is set to $C = \left \{ c_{1},c_{2},\cdots,c_{n}\right \}$, where ${c}_i$ is the output channel number of the $i$-th layer. Channel pruning is to search for an optimal set of channel configurations to maximize the accuracy of the pruned network $\mathcal{N}_{pruned}$ on the validation set $\mathcal{X}_{val}$, provided that the constraints are satisfied.  We can formulate the channel pruning problem as follows:
\begin{align}
\mathop{\max}_{C^{'}}\  & \text{Acc}\left ( \mathcal{N}_{\text{pruned}}\left ( C^{'},\bm{W}^{'} \right );\mathcal{X}_{\text{val}} \right ), \\
\textrm{s.t.} \  & \mathcal{C}(\mathcal{N}_{\text{pruned}}(C^{'},\bm{W}^{'})) < \text{constraint}. \nonumber
\end{align}
Here, $C^{'}=\{ c_{1}^{'},c_{2}^{'},\cdots,c_{n}^{'}\}$ is the channel number configuration of the pruned network subject to $0 < c_{i}^{'} \leq c_{i}$. $\bm{W}^{'}$ is the initialized weight parameters of the pruned network, which can be inherited from the original network or randomly initialized or generated by other ways. $\mathcal{C}$ is the approach for calculating resource constraints, such as FLOPs and parameters.

\subsection{Block Importance}
The recently popular AutoML-based structured pruning methods regards channel pruning as automatic channel search. The size of the search space is $\prod_{i=1}^{n}c_{i}$, which is an enormous number for modern deep neural networks and cannot be solved exhaustively. So researchers have proposed various methods \cite{he2018amc,liu2019metapruning,lin2020channel} to compress the search space and speed up the solution. Reinforcement learning or evolutionary algorithms usually require a lot of time and resources to search for the number of channels in each layer to obtain a relatively optimal structure. In contrast to them, we conceive that the importance of different blocks can be reflected by the mean of their BN layer's scaling factors.

Batch Normalization \cite{ioffe2015batch} is one of the essential components of modern convolutional neural networks and is used to normalize the output of the previous layer to adjust the range, thus helping the network to converge faster and more stably. More specifically, let $X$ and $Y$ be the input and output of a BN layer, BN layer performs the following transformation:
\begin{align}
Y=\gamma \frac{X-\mu }{\sqrt{\sigma ^{2}+\epsilon }}+\beta,
\end{align}
where $\mu$ and $\sigma$ are the mean and standard deviation of the current mini-batch input, $\gamma$ and $\beta$ are the trainable transformation parameters in the BN layer for scaling and shifting, respectively.

In order to emphasize each block's importance in the overall network, following Network Slimming \cite{liu2017learning}, we add L1 regularization to the $\gamma$ parameters of the BN layers in the overall loss function for sparse training. The total loss function is given by the following equation:
\begin{align}
\mathcal{L}=\mathcal{L}_{\text{cls}}(\hat{o},o)+\lambda \sum_{\gamma \in \Gamma }| \gamma |,
\end{align}
where $\hat{o}$ is the prediction result of the model, $o$ is the label, $\mathcal{L}_{\text{cls}}$ is the original loss function of the network, $\lambda$ is used to control the balance of training and sparsity, and $\Gamma$ is the collection of all gamma parameters to be sparse. During the training process, the accuracy of the model and the sparsity of the BN layers gradually reach a balance. 

Block is the basic unit of our pruning method, and the meaning of block is slightly different for varying structure of CNNs. Figure \ref{fig2} illustrates the composition of blocks in different network architectures. For a straight CNN like VGG \cite{simonyan2014very}, the block is the layer. ResNet \cite{he2016deep} naturally takes a residual block as the block to be pruned, while in MobileNetv2 \cite{sandler2018mobilenetv2} it is the inverted residual block. Suppose there are $L$ blocks to be pruned in the current network, and the $i$-th block has a total of $L_{i}$ $\gamma$ parameters that need to be pruned whose mean value is $M_{i}=\frac{1}{L_{i}}\sum_{j=1}^{L_{i}}|\gamma_{j}|$. Then the importance of the block can be measured as:
\begin{align}
I_{i}=\frac{M_{i}}{\sum_{j=1}^{L}M_{j}}.
\end{align}

Network Slimming \cite{liu2017learning} sorts all the $\gamma$ parameters and then prunes the convolutional kernels corresponding to channels with $\gamma$ smaller than the set threshold. This pruning method has two main drawbacks:
\begin{itemize}
\item The threshold is set primarily according to manual experience, making it difficult to obtain a compact sub-network that precisely satisfies the constraints.
\item When the pruning rate is small, those channels with $\gamma$ parameters close to 0 can be easily pruned. However, when the pruning rate is large, this method is no longer reliable and there is a possibility that some layers are overpruned.
\end{itemize}

In contrast, our method can accurately meet the budget constraints and adaptively determine the pruning rate of each layer without over-pruning the important layers and damaging the final accuracy.

\subsection{Bisection Method}
After obtaining the importance evaluation metrics for each block, an obvious question is how to map the importance to the actual pruning rate of each block. Here, we empirically assume that the actual pruning rate of each block is proportional to its importance. Suppose the importance of the $i$-th block is evaluated as $I_{i}$, the pruning rate is $P_{i}$ and the proportion of remaining channels is $R_{i}$, then we have the following equation:
\begin{align}
R_{i} = 1-P_{i}=\alpha I_{i},\ i=1,2,\cdots,L.
\end{align}
Since there may be some unpruned layers in the network, and common constraints such as FLOPs and parameters are closely related to the channel number of the preceding and subsequent layers, it is not easy to solve the value of $\alpha$ directly.

We use the bisection method to find the approximate solution of $\alpha$ expeditiously. Since $\alpha$ controls the final FLOPs assigned to each block, the larger $\alpha$ is, the larger the final FLOPs assigned to each block is. Therefore the variation between the overall FLOPs of the pruned network and $\alpha$ is monotonically increasing. Benefiting from this property, we can gradually reduce the solution interval in the process of bisection search. The first step is to select a wide interval for $\alpha$ to cover the real solution. Then we use the median of the interval to calculate the pruning rate of each block to construct a compact network. Compute the computational resources of the pruned network and determine whether it meets the target budget. If not, update the left and right values of the interval according to the characteristics of bisection method and iterate to solve the problem.
Detailed algorithm is described in Algorithm~\ref{alg:algorithm1}.

For automatic channel pruning methods like ABCPruner \cite{lin2020channel} that search the pruning rate of each layer by an intelligent algorithm, it is usually required to make restrictions on the minimum pruning rate of each layer to reach the pruning budget, which leads to some defects:
\begin{itemize}
\item Limit the range of pruning rates for each layer, which may lead to over-pruning of some layers and make the performance sub-optimal.
\item The minimum pruning rate for each layer is chosen manually by experience and cannot be effectively integrated with the automated pruning process proposed by its framework. Meanwhile, it is hard to precisely reach the pruning budget and is not suitable for some demanding applications.
\item When changing the target budget, the channel configuration needs to be searched again, consuming a lot of computational resources and time.
\end{itemize}

\begin{algorithm}[tb]
\caption{Bisection Method}
\label{alg:algorithm1}
\textbf{Input}: Block Importance metric:\ $I$, Channel Number Configuration:\ $C$, Constraint:\ $\mathcal{C}_{l}$\\
\textbf{Parameter}: Interval Left Value:\ $a$, Interval Right Value:\ $b$,\ Error Tolerance:\ $\delta$\\
\textbf{Output}: Proportionality Factor:\ $\alpha$
\begin{algorithmic}[1] 
\STATE $f(a)=\mathcal{C}(\mathcal{N}_{\text{pruned}}(aI \cdot C))-\mathcal{C}_{l},f(b)=\mathcal{C}(\mathcal{N}_{\text{pruned}}(bI \cdot C))-\mathcal{C}_{l}$.
\WHILE{$a<=b$}
\STATE $\alpha=(a+b)/2,f(\alpha)=\mathcal{C}(\mathcal{N}_{\text{pruned}}(\alpha I \cdot C))-\mathcal{C}_{l}$.
\IF {$|f(\alpha)|<\delta$}
\STATE \textbf{break while}
\ENDIF
\IF {$f(a)\cdot f(\alpha)<0$}
\STATE $b=\alpha,f(b)=f(\alpha)$.
\ELSE
\STATE $a=\alpha,f(a)=f(\alpha)$.
\ENDIF
\ENDWHILE
\STATE \textbf{return} $\alpha$
\end{algorithmic}
\end{algorithm}

\subsection{Adaptive Weights Inheritance}
Through BN sparse training, block importance evaluation, and bisection method search, we can quickly find the compact sub-network that meet the pruning budget. Next, whether to train the sub-network from scratch or to inherit the weights from the large network and fine-tune will be considered. Although it was shown in a previous study through extensive experiments that the pruned sub-network can achieve close to or even better performance with the same training resources as the large network without inheriting the weights from the original network (i.e., random initialization). However, it is found in our experiments that the inherited weights of the large network can help the sub-network converge to higher accuracy without investing more training resources, using a carefully set learning rate and more data augmentation. In other words, proper selection of weights of large network to initialize the sub-network can achieve better performance performance with limited training resources.

MetaPruning \cite{liu2019metapruning} first recalibrates the mean $\mu$ and standard deviation $\sigma$ of the BN layer with a small amount of data during the search of the sub-networks, and then directly evaluates their accuracies on the validation set. This is because some channels in the network are discarded after pruning, and the mean $\mu$ and standard deviation $\sigma$ statistics of the activations are changed accordingly, then using the old values naturally causes a mismatch. It is also shown experimentally in EagleEye \cite{li2020eagleeye} that sub-networks with recalibrated BN parameters maintain a high degree of performance consistency before and after training. Specifically, if a sub-network after recalibrating the BN parameters performs well on the validation set, then it can be assumed that the accuracy is also good after full training. The recalibration of the BN parameters is done by fixing all trainable parameters of the model and using a very small number (probably a few thousand) of training set images to inference network only, and updating the mean $\mu$ and standard deviation $\sigma$ of the BN layer.

Motivated by the above work, we will quickly evaluate the performance of the sub-networks according to different weight inheritance criteria in the same way. Taking a layer in the network as an example, suppose its output channel number is $C_{o}$. Given its pruning rate $P_{i}$, we choose different weight inheritance criteria to initialize the sub-network. Our work evaluated three kinds of metrics:
\begin{itemize}
\item $l_{1}$-norm \quad Calculate the $l_{1}$-norm of convolution kernel weights corresponding to the output channels and sort them in descending order, then inherit the weights of the first $C_{o} \cdot (1-P_{i})$ channels.
\item BN weights \quad After BN sparse training, the absolute values of $\gamma$ corresponding to different channels reflect the importance of that channel. We rank the absolute values of all scaling factors in descending order and inherit the weights of the first $C_{o} \cdot (1-P_{i})$ channels.
\item Geometric Median \quad FPGM \cite{he2019filter} views convolutional filters as points in Euclidean spaces, and since the geometric median is a classic robust estimator of centrality for data, if a filter is close to the GM, it is considered that its information can be represented by other filters and thus can be pruned. We prune the $C_{o} \cdot P_{i}$ filters closest to the GM and inherit the weights of the remaining channels.
\end{itemize}

After recalibrating the BN parameters separately for the sub-networks with different weight inheritance criteria, we first evaluate their accuracies on the validation set, and then select the sub-network with the highest accuracy for subsequent fine-tuning. In addition, various existing pruning techniques for evaluating channel importance can be used in our pruning framework, arguably providing a more broadly general platform for channel pruning.

\section{Experiments}
In this section, we demonstrate the effectiveness of our proposed method on CIFAR-10 and ImageNet datasets for different network architectures. Experimental details are first presented, then we compare with other SOTA channel pruning methods and finally, ablation experiments are implemented to show the effectiveness of adaptive channel pruning and adaptive weight inheritance, respectively.

\begin{table}[t]
\centering
\resizebox{\columnwidth}{!}{
    \begin{tabular}{c|c|c|c}
    \toprule
    Model   & Method & $\Delta$FLOPs & Acc(\%) \\
    \midrule
    \multirow{8}[2]{*}{{VGG16}}
        & L1 norm   & -34\% & 93.40 \\
        & NetworkSlimming   & -51\% & 93.80 \\
        & ThiNet & -50\% & 93.85 \\
        & CP    & -50\% & 93.67 \\
        & HRank & -53.5\%   & 93.43 \\
        & \textbf{AdaPruner}(Ours) & -50\% & \textbf{94.02} \\
        & ABCPruner & -73\% & 93.08 \\
        & \textbf{AdaPruner}(Ours) & -73\% & \textbf{93.50} \\
    \midrule
    \multirow{8}[2]{*}{{ResNet56}}
        & ThiNet & -50\% & 92.98 \\
        & CP    & -50\% & 92.80 \\
        & SFP   & -52.6\%   & 93.35 \\
        & FPGM  & -52.6\%   & \textbf{93.49} \\
        & AMC  & -50\%   & 91.90 \\
        & HRank & -50\%   & 93.17 \\
        & ABCPruner & -54\% & 93.23 \\
        & \textbf{AdaPruner}(Ours) & -50\% & \textbf{93.49} \\
    \midrule
    \multirow{7}[2]{*}{{ResNet110}}
        & L1 norm   & -40\% & 93.30 \\
        & SFP   & -40\%   & 93.86 \\
        & FPGM  & -52.3\%   & 93.74 \\
        & \textbf{AdaPruner}(Ours) & -50\% & \textbf{93.90} \\
        & HRank & -68.6\%   & 92.65 \\
        & ABCPruner & -65\% & 93.58 \\
        & \textbf{AdaPruner}(Ours) & -65\% & \textbf{93.87} \\
    \bottomrule
    \end{tabular}
    }
    \caption{Comparison of different methods for pruning VGG16, ResNet56, and ResNet110 on the CIFAR-10 dataset.}
\label{table1}
\end{table}

\subsection{Implementation Details}
\subsubsection{Training Strategy.}We conduct experiments on the benchmark CIFAR-10 \cite{krizhevsky2009learning} and ImageNet \cite{russakovsky2015imagenet} datasets for image classification. To compare with other works, we chose the mainstream CNN models. We prune VGG16, ResNet56 and ResNet110 on CIFAR-10, while ResNet50 and MobileNetV2 are pruned on ImageNet.

On CIFAR-10, We refer to the training settings of ABCPruner \cite{lin2020channel}. Using the SGD optimizer, the momentum is set to 0.9, the weight decay is kept at 5e-3, the initial learning rate is set to 0.01, and every 50 epochs, the learning rate is reduced to one-tenth of the previous one, with a total training of 150 epochs and batch size of 256.

On ImageNet, the training settings for ResNet50 are consistent with \cite{lin2020channel}. The weight decay is set to 1e-4, the learning rate is 0.1, and divided by 10 every 30 epochs for a total of 90 epochs. For MobileNetV2, we use the SGD optimizer with momentum set to 0.9 and weight decay to 5e-4. The learning rate is gradually decayed from 0.05 to 0 via the cosine schedule. A total training is 150 epochs with the batch size of 512.

Following \cite{liu2017learning}, the $\lambda$ used to sparse the scaling factors of BN layers is set to 0.0001.

\subsubsection{Pruning Setting.} AdaPruner only requires the initial interval and target budget error to be given when pruning the network. Since the search is fast enough, the left and right values of the interval can be relaxed appropriately. In our paper, the left value of the interval is set to 0.01 and the right value to 100. Fine-tuning the pruned network employs the same training strategy as the one used for sparse training.

\begin{table}[t]
\centering
\resizebox{\columnwidth}{!}{
    \begin{tabular}{c|c|c|c|c|c}
    \toprule
    Model   & Method & Baseline(\%) & Pruned(\%) & $\Delta$Acc(\%) & FLOPs \\
    \midrule
    \multirow{8}[2]{*}{\rotatebox{90}{ResNet50}}
        & ThiNet & 72.88 & 71.01  & -1.87 & 1.7G \\
        & AutoPruner & 76.15 & 74.76  & -1.39 & 1.88G \\
        & FPGM & 76.15   & 74.83  & -1.32 & 1.91G \\
        & HRank & 76.15   & 74.98  & -1.17 & 2.3G \\
        & MetaPruning    & 76.6 & 75.4 & -1.2 & 2.0G \\
        & PFS & 77.2 & 75.6  & -1.6 & 2.0G \\
        & AutoSlim   & 76.6   & 75.6 & -1.0 & 2.0G \\
        & EagleEye   & 77.28   & 76.4 & -0.88 & 2.0G \\
        & ABCPruner & 76.01 & 73.86  & -2.15 & 1.89G \\
        & \textbf{AdaPruner} & 76.46 & 75.66  & \textbf{-0.8} & 2.0G \\
    \midrule
    \multirow{8}[2]{*}{\rotatebox{90}{MobileNetV2}}
        & Uniform 1.0x & 72.0 & -  & - & 300M \\
        & Uniform 0.75x & 72.0 & 69.8  & -2.2 & 210M \\
        & GFP & 75.74 & 73.42  & -2.32 & 290M \\
        & AMC & 71.8 & 70.8  & -1.0 & 211M \\
        & MetaPruning    & 72.0 & 71.2 & -0.8 & 217M \\
        & AutoSlim   & 74.2   & 73.0 & -1.2 & 207M \\
        & PFS & 72.1 & 70.9  & -1.2 & 210M \\
        & AACP  & 71.8  & 71.1 & -0.7 & 210M \\
        & GFS  & 72.0  & 71.2 & -0.8 & 201M \\
        & \textbf{AdaPruner} & 71.49 & 70.87 & \textbf{-0.62} & 211M \\
    \bottomrule
    \end{tabular}
    }
    \caption{Comparison of different channel pruning methods for compressing ResNet50 and MobileNetV2 on the ImageNet dataset. Here we report the Top-1 accuracy.}
\label{table2}
\end{table}

\subsection{Results on CIFAR-10}
We prune VGG16, ResNet56 and ResNet110 to varying degrees on the CIFAR-10 dataset and compare their accuracies with previous channel pruning methods, including L1 norm based pruning method \cite{li2016pruning}, Network Slimming \cite{liu2017learning}, ThiNet \cite{luo2017thinet}, Channel Pruning (CP) \cite{he2017channel}, Soft Filter Pruning (SFP) \cite{he2018soft}, FPGM \cite{he2019filter}, AMC \cite{he2018amc}, HRank \cite{lin2020hrank}, and ABCPruner \cite{lin2020channel}. Since the size of the CIFAR-10 dataset is quite small and we further train the pre-trained network with sparsity, which is therefore fully trained, we are more concerned with the accuracy of the pruned model.

The pruning results of the CIFAR-10 dataset are illustrated in Table \ref{table1}. For VGG16, after pruning 50\% FLOPs, the fine-tuned compact network achieves 94.02\% accuracy, which is better than the traditional pruning methods like L1 norm, Network Slimming, CP, HRank and ThiNet. To compare with ABCPruner, we also reduce 73\% FLOPs of VGG16, and the results indicate that AdaPruner is also superior to the AutoML-based pruning method. As for ResNet56, our method reduces the FLOPs by 50\% and the accuracy is on par with that of FPGM, which is better than other pruning methods. When reduced 50\% FLOPs of the ResNet110, the accuracy of the compact network is 93.90\%. As the pruning rate is further increased to 65\%, AdaPruner also has a significantly better pruning result than HRank and ABCPruner, with only a small decrease in accuracy compared to the 50\% pruning rate. This suggests that our pruning method can adaptively assign the FLOPs for each block according to the results of sparse training and the target budget, moreover, our method is once-and-for-all and does not require re-searching each layer's channel number.

\begin{figure}[t]
\centering
\includegraphics[width=0.9\columnwidth]{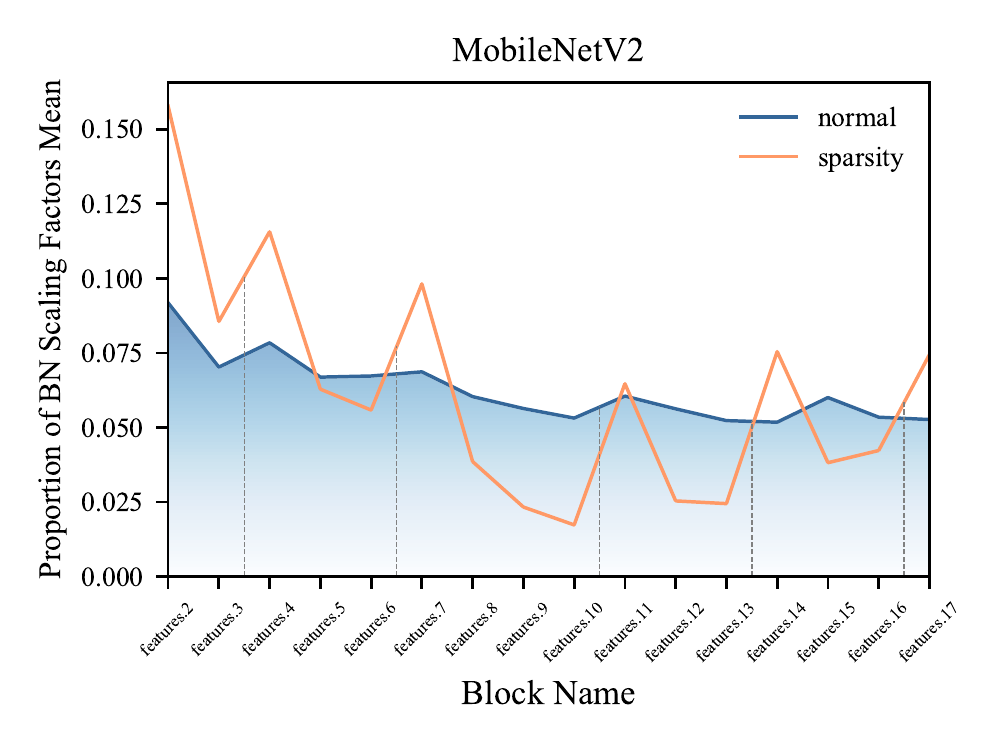} 
\caption{Comparison results of the proportion of BN scaling factors mean for normal training and sparse training in MobileNetV2. Best viewed in color.}
\label{fig3}
\end{figure}

\begin{figure}[t]
\centering
\includegraphics[width=0.9\columnwidth]{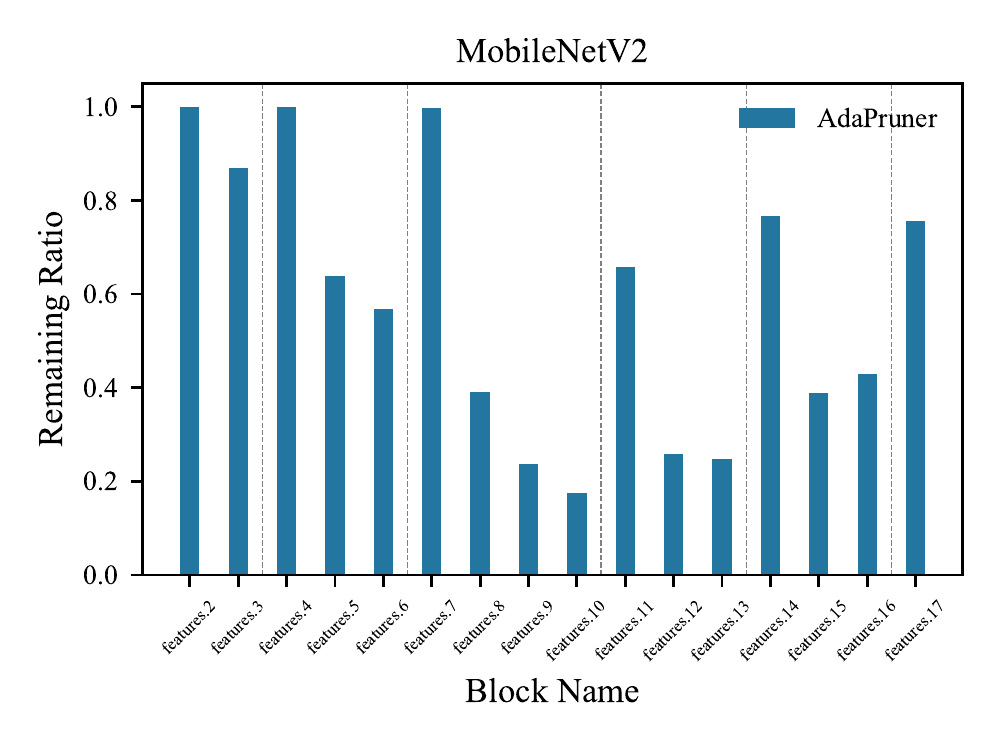} 
\caption{Visualization of the ratio of remained channels for each block of MobileNetV2 pruned by AdaPruner.}
\label{fig4}
\end{figure}

\subsection{Results on ImageNet}
We prune ResNet50 and MobileNetV2 on the ImageNet dataset and compare their accuracies with previous channel pruning methods, including ThiNet \cite{luo2017thinet}, AutoPruner\cite{luo2020autopruner}, FPGM \cite{he2019filter}, HRank \cite{lin2020hrank}, MetaPruning \cite{liu2019metapruning}, Pruning from Scratch (PFS) \cite{wang2020pruning}, AutoSlim \cite{yu2019autoslim}, EagleEye \cite{li2020eagleeye}, ABCPruner \cite{lin2020channel}, Group Fisher Pruning (GFP) \cite{liu2021group}, Greedy Forward Selection (GFS) \cite{ye2020good} and AACP \cite{lin2021aacp}.

Table \ref{table2} shows the results of different channel pruning methods on the ImageNet. Our approach exhibits excellent performance when ResNet50 reduces the FLOPs by more than 50\%, meaning that the amount of computational operations in the pruned network remains around 2G. Compared with the baseline accuracy of the unpruned model, the Top-1 accuracy of the compact network obtained by AdaPruner decreases by only 0.8\%, which is significantly better than traditional pruning methods such as ThiNet, AutoPruner, FPGM and HRank. Compared with the current state-of-the-art AutoML and NAS based pruning methods like MetaPruning, AutoSlim, EagleEye and ABCPruner, our method does not require extensive search but still achieves quite competitive results. Efficient compression of lightweight networks like MobileNetV2 is currently a challenge in the field of model pruning. AdaPruner gives a concise and efficient solution. With 30\% reduction of FLOPs, the accuracy of our compact model drops only 0.62\% compared to the baseline, which exceeds previous state-of-the-art methods to our knowledge.

\begin{figure}[t]
\centering
\includegraphics[width=0.9\columnwidth]{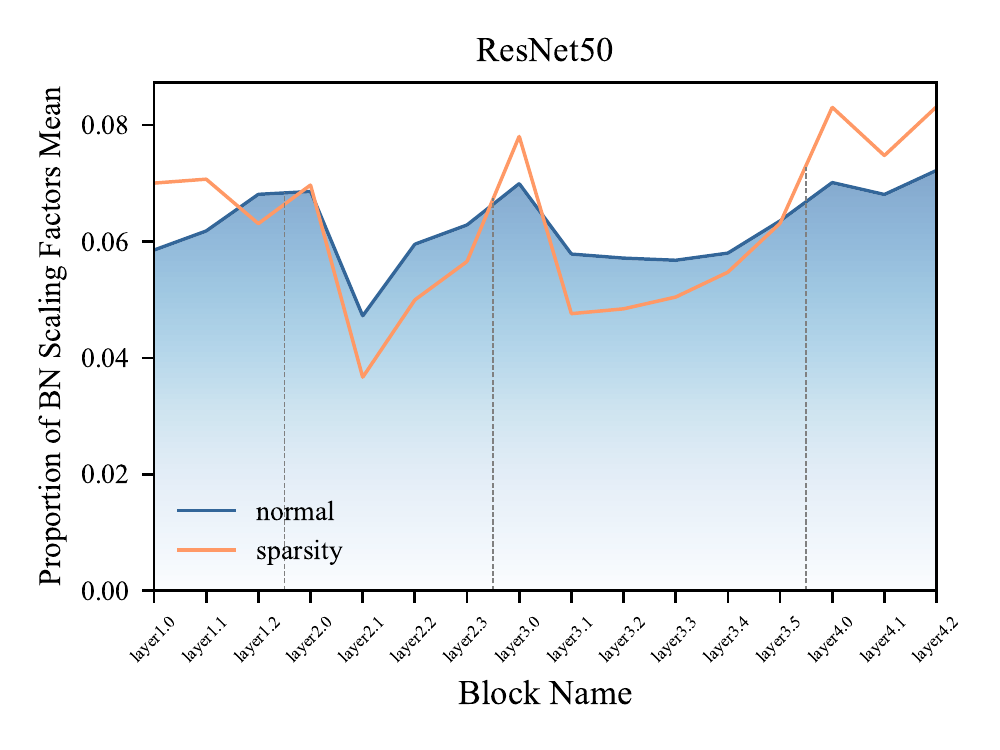} 
\caption{Comparison results of the proportion of BN scaling factors mean for normal training and sparse training in ResNet50. Best viewed in color.}
\label{fig5}
\end{figure}

\begin{figure}[t]
\centering
\includegraphics[width=0.9\columnwidth]{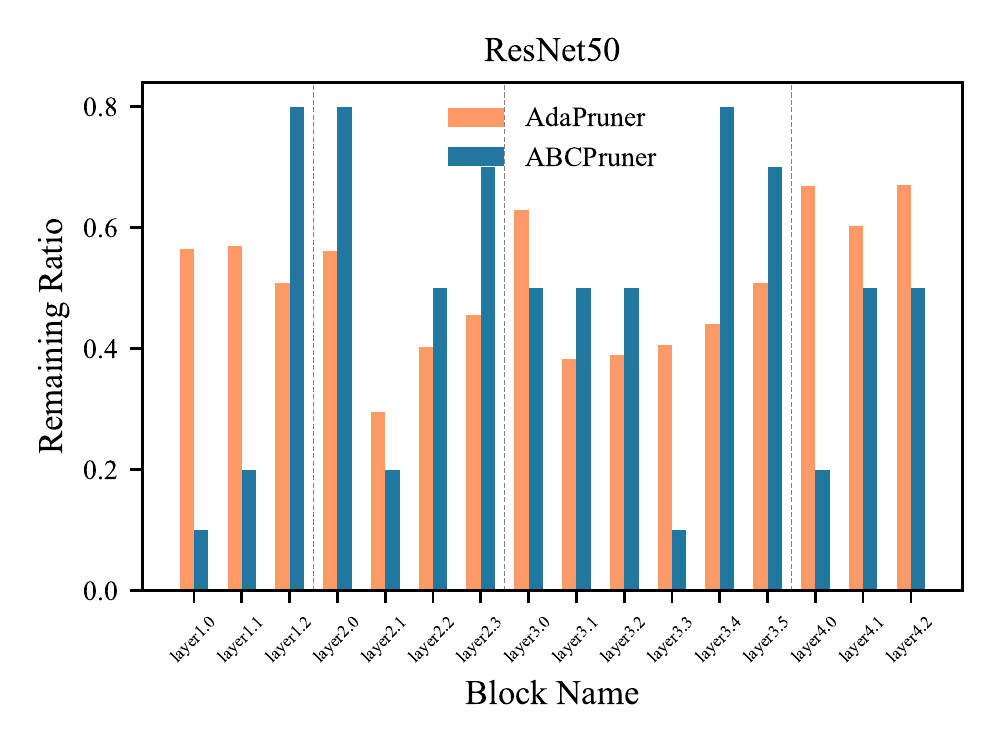} 
\caption{Visualization of the ratio of remained channels for each block of ResNet50 pruned by AdaPruner and ABCPruner. Best viewed in color.}
\label{fig6}
\end{figure}

\subsection{Pruning Result Visualization}
In order to observe the pruning of our proposed method for networks with different structures, in this part we visualize the pruning results of AdaPruner on ResNet50 and MobileNetV2. As the pruning rate of each block is closely related to its importance, we also plot the change of the proportion of each block's BN scaling factors mean before and after training with sparsity.

\subsubsection{Proportion of BN Scaling Factors Mean.} Figure \ref{fig3} and Figure \ref{fig5} show the comparison of the proportion of each block's BN scaling factors mean (i.e., Block Importance $I_{i}$ mentioned above) prior to and after sparse training for MobileNetV2 and ResNet50, respectively. For MobileNetV2, the sparse training about the BN layer significantly amplifies the importances of the different blocks. In order to ensure the accuracy, the first two convolutional layers are not pruned, and the rest can be divided into six stages according to the output channel number, which are distinguished by dashed lines in Figure \ref{fig5}. It is not hard to find that the importance of each stage's first block is the highest in this stage, and we guess it is because the first block of each stage is used for downsampling and changing the output channel number, carrying the role of encoding more information and thus cannot be over-pruned. In fact, a similar phenomenon is found in the experiments of MetaPruning \cite{liu2019metapruning}. As for ResNet50, similar results can be observed and the blocks of the last stage have higher importances as they contribute to image classification.

\subsubsection{Channel Remaining Ratio.} The pruning result of MobileNetV2 is shown in Figure \ref{fig4}. As we can see, our pruning approach first allocates more resources in the first block of each stage, and then tends to keep more channels in the first two stages and prunes significantly in the middle layers. Figure \ref{fig6} shows the proportion of different blocks' remaining channels in the pruned ResNet50 using our proposed methods AdaPruner and ABCPruner \cite{lin2020channel}, respectively, with the same budget conditions. In our approach, the pruning result of ResNet50 reflect similar properties to that of MobileNetV2. More channels are reserved for the initial block of each stage, with a tendency to prune the intermediate stages and keep more resources in the final stage. In contrast, the pattern of pruning results of ABCPruner is not obvious, which may be attributed to the inadequate convergence of the ABC algorithm on large dataset.

\begin{figure}[t]
\includegraphics[width=1.0\columnwidth]{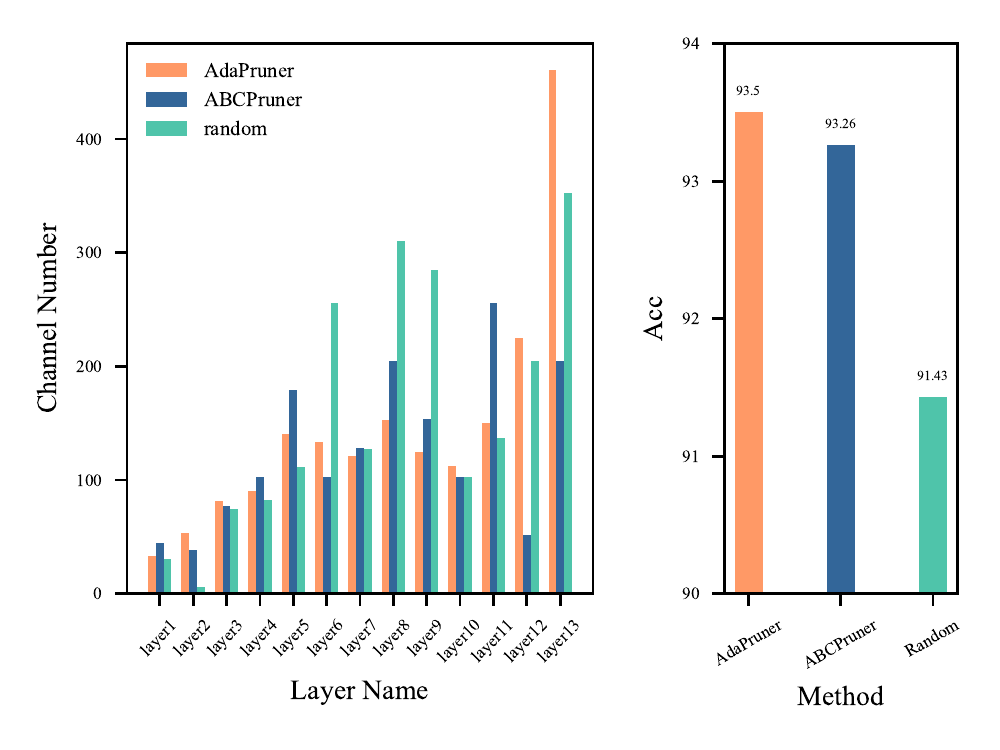} 
\caption{Comparison of pruning results and fine-tuned accuracy of AdaPruner, ABCPruner and random methods on VGG16. Best viewed in color.}
\label{fig7}
\end{figure}

\subsection{Ablation Study}
Our pruning method integrates the architecture and initialization weights of the pruned sub-network. In this part, we will demonstrate that the sub-network architecture and weights obtained by AdaPruner are both good enough through ablation experiments.
\subsubsection{Pruned Network Architecture.} To demonstrate that our method of assigning pruning rates to each layer by the proportion of BN scaling factors mean is superior to the method of searching by AutoML, we prune VGG16 according to the pruning rates obtained by AdaPruner, ABCPruner and random sampling, respectively, under the same constraints (i.e., 50\% reduction of FLOPs), and all are guaranteed to use the $l_{1}$-norm weight inheritance criterion to train. The random seed is kept consistent for each experiment for fair comparison. Figure \ref{fig7} compares the retained number of channels and the fine-tuned accuracies of different pruning methods. Random pruning is prone to over-pruning because it does not take into account the importance of the different layers, which makes the accuracy lower. ABCPruner uses an intelligent optimization algorithm to find a suitable sub-network architecture, but the convergence process takes too long and tends to fall into sub-optimal solutions. In contrast, our method adaptively determines the importance of each layer during the sparse training and thus achieves the best accuracy.

\begin{table}[t]
\centering
\resizebox{\columnwidth}{!}{
    \begin{tabular}{c|c|c|c|c}
    \toprule
    \multirow{2}{*}{Acc(\%)} & \multicolumn{2}{c|}{VGG16}      & \multicolumn{2}{c}{ResNet56}   \\ \cline{2-5}
     & Re             & Final          & Re             & Final          \\
    \midrule
    random                     & 8.75           & 93.56          & 9.99           & 92.03          \\
    \midrule
    $l_{1}$-norm                  & \textbf{42.84} & \textbf{94.02} & \textbf{80.32} & \textbf{93.49} \\
    \midrule
    BN weights               & 36.52          & 93.91          & 77.3           & 93.14          \\
    \midrule
    GM                       & 41.79          & 93.88          & 78.08          & 93.27          \\
    \bottomrule
    \end{tabular}
}
\caption{Comparison of the final accuracies of different initialization weights with the same sub-network architecture. Random represents the random initialization weights, not inheriting the weights of the original network. GM stands for inheriting the weights of the original network according to Geometric Median criterion. Re denotes the accuracy of the model in the validation set by recalibrating the BN parameters. Final represents the accuracy after complete training.}
\label{table3}
\end{table}

\subsubsection{Weight Inheritance Criteria.} For comparing the effect of different initialization weights on the final accuracy, we prune VGG16 and ResNet56 according to the pruning rates obtained from AdaPruner under the same constraints (i.e., 50\% reduction of FLOPs), and then initialize them using four pre-designed weight inheritance criteria, which are random initialization, $l_{1}$-norm, BN weights, and Geometric Median. The random seed is kept consistent for each experiment for fair comparison. From the experimental results in Table \ref{table3}, we can draw the following conclusions: (1) With limited training resources, the initial weights have a significant impact on the accuracy of the pruned model. (2) The accuracy of the network after recalibration of BN parameters is highly correlated with that following full training. (3) Our method can automatically select the weight inheritance criterion that best suits current network architecture and dataset, resulting in high accuracy. It is worth mentioning that we found that on the CIFAR-10 dataset, the models tend to choose either $l_{1}$-norm or GM. On the large ImageNet dataset, all models coincidentally choose GM as the weight inheritance criterion.

\section{Conclusion}
In this paper, we propose a channel pruning method that integrates the network architecture and initialization weights, named AdaPruner. Our method determines the importance of each block based on the proportion of each block's BN scaling factors mean after training with sparsity, and quickly finds the pruning rate corresponding to each block under the budgetary condition by bisection method, avoiding the extensive search of previous methods. The weight inheritance criterion that best fits the current architecture is adaptively selected for initialization, helping the pruned network to converge to high accuracy. Our pruning framework works in conjunction with many previous pruning methods, providing the community with a widely applicable pruning pipeline.

\bibliography{aaai22}

\end{document}